\documentclass[11pt,a4paper]{article}
\usepackage[hyperref]{emnlp2018}
\usepackage{times}
\usepackage{latexsym}
\usepackage{graphicx}
\usepackage{tabularx}
\usepackage{anyfontsize}
\usepackage{mathtools}
\usepackage{amsmath}
\usepackage{amssymb}
\usepackage[flushleft]{threeparttable}

\usepackage{url}

\aclfinalcopy 

\newcommand{\R}{\mathbb{R}}

\title{Paraphrase Detection on Noisy Subtitles in Six Languages}

\author{Eetu Sj\"oblom
  \\\And
  Mathias Creutz \vspace{1mm} \\
 {\normalsize Department of Digital Humanities, Faculty of Arts } \\
  University of Helsinki \\
  {\small Unioninkatu 40, FI-00014 University of Helsinki, Finland} \\
 {\small \tt \{eetu.sjoblom, mathias.creutz, mikko.aulamo\}@helsinki.fi}
 \\\And
  Mikko Aulamo }

\date{}

\begin{document}
\maketitle


\begin{abstract}
We perform automatic paraphrase detection on subtitle data from the Opusparcus corpus comprising six European languages: German, English, Finnish, French, Russian, and Swedish. We train two types of supervised sentence embedding models: a word-averaging (WA) model and a gated recurrent averaging network (GRAN) model. We find out that GRAN outperforms WA and is more robust to noisy training data. Better results are obtained with more and noisier data than less and cleaner data. Additionally, we experiment on other datasets, without reaching the same level of performance, because of domain mismatch between training and test data.
\end{abstract}

\section{Introduction}

This paper studies automatic paraphrase detection on subtitle data for six European languages. Paraphrases are a set of phrases or full sentences in the same language that mean approximately the same thing. Automatically finding out when two phrases mean the same thing is interesting from both a theoretical and practical perspective. Theoretically, within the field of distributional, compositional semantics, there is currently a significant amount of interest in models and representations that capture the meaning of not just single words, but sequences of words. There are also practical implementations, such as providing multiple alternative correct translations when evaluating the accuracy of machine translation systems.

To our knowledge, the present work is the first published study of automatic paraphrase detection based on data from \textit{Opusparcus}, a recently published paraphrase corpus \cite{creutz2018lrec}\footnote{Opusparcus is available for download at: \url{http://urn.fi/urn:nbn:fi:lb-201804191}}. Opusparcus consists of \textit{sentential} paraphrases, that is, pairs of full sentences that convey approximately the same meaning. Opusparcus provides data for six European languages: German, English, Finnish, French, Russian, and Swedish. The data sets have been extracted from OpenSubtitles2016 \cite{lison2016lrec}, which is a collection of translated movie and TV subtitles.\footnote{OpenSubtitles2016 is extracted from \url{www.opensubtitles.org}. OpenSubtitles2016 is in itself a subset of the larger OPUS collection (``... the open parallel corpus''): \url{opus.lingfil.uu.se}, and provides a large number of sentence-aligned parallel corpora in 65 languages.}

In addition to Opusparcus, experiments are performed on other well known paraphrase resources: (1)~\textit{PPDB}, the Paraphrase Database \cite{ganitkevitch2013naacl,ganitkevitch2014lrec,pavlick2015acl}, (2)~\textit{MSRPC}, the Microsoft Research Paraphrase Corpus \cite{quirk2004emnlp,dolan2004coling,dolan2005iwp}, (3)~\textit{SICK} \cite{marelli2014lrec}, and (4)~\textit{STS14} \cite{agirre2014semeval}.

We are interested in movie and TV subtitles because of their conversational nature. This makes subtitle data ideal for exploring dialogue phenomena and properties of everyday, colloquial  language \cite{paetzold2016coling,vanderwees2016coling,lison2018lrec}. We would also like to stress the importance of working on other languages beside English. Unfortunately, many language resources contain English data only, such as MSRPC and SICK. In other datasets, the quality of the English data surpasses that of the other languages to a considerable extent, as in the mutilingual version of PPDB \cite{ganitkevitch2014lrec}.

Although our subtitle data is very interesting data, it is also noisy data, in several respects. Since the subtitles are \textit{user-contributed data}, there are misspellings both due to human mistake and due to errors in optical character recognition (OCR). OCR errors emerge when textual subtitle files are produced by ``ripping'' (scanning) the subtitle text from DVDs using various tools. Furthermore, movies are sometimes not tagged with the correct language, they are encoded in various character encodings, and they come in various formats. \cite{tiedemann2007clin,tiedemann2008lrec,tiedemann2016lrec}

A different type of errors emerge because of misalignments and issues with sentence segmentation. Opusparcus has been constructed by finding pairs of sentences in one language that have a common translation in at least one other language. For example, English \textit{``Have a seat.''} is potentially a paraphrase of \textit{``Sit down.''} because both can be translated to French \textit{``Asseyez-vous.''} \cite{creutz2018lrec} To figure out that \textit{``Have a seat.''} is a translation of  \textit{``Asseyez-vous.''}, English and French subtitles for the same movie can be used. English and French text that occur at the same time in the movie are assumed to be translations of each other. However, there are many complications involved: Subtitles are not necessarily shown as entire sentences, but as snippets of text that fit on the screen. There are numerous partial overlaps when comparing the contents of subtitle screens across different languages, and the reconstruction of proper sentences may be difficult. There may also be timing differences, because of different subtitle speeds and different time offsets for starting the subtitles. \cite{tiedemann2007clin,tiedemann2008lrec} Furthermore, \newcite{lison2018lrec} argue that ``[subtitles] should better be viewed as boiled down transcriptions of the same conversations across several languages. Subtitles will inevitably differ in how they `compress' the conversations, notably due to structural divergences between languages, cultural differences and disparities in subtitling traditions/conventions. As a consequence, sentence alignments extracted from subtitles often have a higher degree of insertions and deletions compared to alignments derived from other sources.''

We tackle the paraphrase detection task using a sentence embedding approach. We experiment with sentence encoding models that take as input a single sentence and produce a vector representing the semantics of the sentence. While models that rely on sentence pairs as input are able to use additional information, such as attention between the sentences, the sentence embedding approach has its advantages: Embeddings can be calculated also when no sentence pair is available, and large numbers of embeddings can be precalculated, which allows for fast comparisons in huge datasets.

Sentence representation learning has been a topic of growing interest recently. Much of this work has been done in the context of general-purpose sentence embeddings using unsupervised approaches inspired by work on word embeddings \cite{hill2016naacl, kiros2015nips} as well as approaches relying on supervised training objectives \cite{conneau2017emnlp, subramanian2018arxiv}. While the paraphrase detection task is potentially useful for learning general purpose embeddings, we are mainly interested in paraphrastic sentence embeddings for paraphrase detection and semantic similarity tasks.

Closest to the present work is that of \citet{wieting2017acl}, who study sentence representation learning using multiple encoding architectures and two different sources of training data. It was found that certain models benefit significantly from using full sentences (SimpWiki) instead of short phrases (PPDB) as training data. However, the SimpWiki data set is relatively small, and this leaves open the question how much the approaches could benefit from very large corpora of sentential paraphrases. It is also unclear how well the approaches generalize to languages other than English.

The current paper takes a step forward in that experiments are performed on five other languages in addition to English. We also study the effects of noise in the training data sets.

\section{Data}

Opusparcus \cite{creutz2018lrec} contains so-called training, development and test sets for each of the six languages it covers. The training sets, which consist of millions of sentence pairs, have been created automatically and are orders of magnitude larger than the development and test sets, which have been annotated manually and consist of a few thousands of sentence pairs. The development and test sets have different purposes, but otherwise they have identical properties: the development sets can be used for optimization and extensive experimentation, whereas the test sets should only be used in final evaluations.

The development and test sets are ``clean'' (in principle), since they have been checked by human annotators. The annotators were shown pairs of sentences, and they needed to decide whether the two sentences were paraphrases (that is, meant the same thing), on a four-grade scale: \textit{dark green} (good), \textit{light green} (mostly good), \textit{yellow} (mostly bad), or \textit{red} (bad). Two different annotators checked the same sentence pairs and if the annotators were in full agreement or if they chose different but adjacent categories, the sentence pair was included in the data set. Otherwise the sentence pair was discarded.

There was an additional choice for the annotators to explicitly discard bad data. Data was to be discarded, if there were spelling mistakes, bad grammar, bad sentence segmentation, or the language of the sentences was wrong. The highest ``trash rate'' of around 11\,\% occurred for the French data, apparently because of numerous grammatical mistakes in French spelling, which is known to be tricky. The lowest ``trash rate'' of below 3\,\% occurred for Finnish, a language with highly regular orthography. Interestingly, English was second best after Finnish, with less than 4\,\% discarded sentence pairs. Although English orthography is not straightforward, there are few diacritics that can go wrong (such as accents on vowels), and English benefits from the largest amounts of data and the best preprocessing tools. Table~\ref{tab:trash} displays a breakdown of the error types in the English and Finnish annotated data.

\begin{table}[!h]
  \begin{center}
    \begin{tabular}{| l | r r | r r |}
      \hline
      Type & \multicolumn{2}{c|}{English} & \multicolumn{2}{c|}{Finnish} \\ \hline
      Not grammatical & 64 & \small (54\%)  & 35 & \small (36\%) \\
      OCR error & 13 & \small (11\%) & 22 & \small (23\%) \\
      Wrong language & 28 & \small (24\%) & 12 & \small (13\%)\\
      Actually correct & 14 & \small (12\%) & 27 & \small (28\%) \\
      \hline
      Total & 119 & \small (100\%) & 96 & \small (100\%) \\
      \hline
    \end{tabular}
  \end{center}
  \caption{\label{tab:trash} The numbers and proportions of different error types in the data discarded by the annotators. Note that some of the sentence pairs that have been discarded are actually correct and have been mistakenly removed by the annotators.}
\end{table}

The Opusparcus training sets need to be much larger than the development and test sets in order to be useful. However, size comes at the expense of quality, and the training sets have not been checked manually. The training sets are assumed to contain noise to the same extent as the development and test sets. On one hand, when it comes to spelling and OCR errors, this may not be too bad, as a paraphrase detection model that is robust to noise is a good thing. On the other hand, when we train a supervised paraphrase detection model, we would like to know which of the sentence pairs in the training data are actual paraphrases and which ones are not. Since the training data has not been manually annotated, we cannot be sure. Instead we need to rely on the automatic ranking presented by \newcite{creutz2018lrec} that is supposed to place the sentence pairs that are most likely to be true paraphrases first in the training set and the sentence pairs that are least likely to be paraphrases last.

In the current paper, we investigate whether it is more beneficial to use less and cleaner training data or more and noisier training data. We also compare different models in terms of their robustness to noise.

In addition to the Opusparcus data, we use other data sources. In Section~\ref{sec:ppdb} we experiment with a model trained on PPDB, a large collection of noisy, automatically extracted and ranked paraphrase candidates. PPDB has been successfully used in paraphrase models before \cite{wieting2015tacl, wieting2016iclr, wieting2017acl}, so we are interested in comparing the performance of models trained on Opusparcus and those trained on PPDB.

We also evaluate our models on MSRPC, a well-known paraphrase corpus. While Opusparcus contains mostly short sentences of conversational nature, and PPDB contains mostly short phrases and sentence fragments, the MSRPC data comes from the news domain. MSRPC was created by automatically extracting potential paraphrase candidates, which were then checked by human annotators.

Lastly, two semantic textual similarity data sets, SICK and STS14 are used for evaluation in a transfer learning setting. SICK contains sentence pairs from image captions and video descriptions annotated for relatedness with scores in the \([0, 5]\) range. It consists of about 10,000 English sentences which are descriptive in nature. STS14 comprises five different subsets, ranging over multiple genres, also with human-annotated scores within \([0, 5]\).

\section{Embedding models}

We use supervised training to produce sentence embedding models, which can be used to determine how similar sentences are semantically and thus if they are likely to be paraphrases.

\subsection{Models}

In our models, there is a sequence of words (or subword units) to be embedded: \(s = (w_1, w_2, ..., w_n)\). The embedding of a sequence \(s\) is \(g(s)\), where \(g\) is the embedding function.

The word embedding matrix is \(W \in \R^{d \times |V|}\), where \(d\) is the dimensionality of the embeddings and \(|V|\) is the size of the vocabulary. \(W^{w_i}\) is used to denote the embedding for the token \(w_i\).

We use a simple word averaging (WA) model as a baseline. In this model the phrase is embedded by averaging the embeddings of its tokens:
    \[g(s) = \frac{1}{n} \sum_{i=1}^{n} W^{w_i}\]

\noindent Despite its simplicity, the WA model has been shown to achieve good results in a wide range of semantic textual similarity tasks. \cite{wieting2016iclr}

Our second model is a variant of the gated recurrent averaging network (GRAN) introduced by \citet{wieting2017acl}. GRAN extends the WA model with a recurrent neural network, which is used to compute gates for each word embedding before averaging. We use a gated recurrent unit (GRU) network \cite{cho2014ssst}. The hidden states \((h_1, ..., h_n)\) are computed using the following equations:
\begin{equation*}
\begin{split}
    r_t &= \sigma(W_r W^{w_t} + U_r h_{t-1}) \\
    z_t &= \sigma(W_z W^{w_t} + U_z h_{t-1}) \\
    \tilde{h}_t &= z_t \circ f(W_h W_{w_t} + U_h(r_t \circ h_{t-1}) + b_h) \\
    h_t &= (1 - z_t) \circ h_{t - 1} + \tilde{h}_t\\
\end{split}
\end{equation*}

 \noindent Here \(W_r\), \(W_z\), \(W_h\), \(U_r\), \(U_z\), and \(U_h\) are the weight matrices, \(b_h\) is a bias vector, \(\sigma\) is the sigmoid function, and \(\circ\) denotes the element-wise product of two vectors.

At each time step \(t\) we compute a gate for the word embedding and elementwise-multiply the gate with the word embedding to acquire the new word vector \(a_t\):
\begin{align*}
    &g_t = \sigma(W_x W^{w_t} + W_h h_t + b) \\
    &a_t = W^{w_t} \circ g_t
\end{align*}

\noindent Here \(W_x\) and \(W_h\) are weight matrices. The final sentence embedding is computed by averaging the word vectors:
    \[g(s) = \frac{1}{n} \sum_{i=1}^{n} a_i\]

%
%
%

\subsection{Training}
\label{sec:training}

Our training data consists of pairs of sequences \((s_1, s_2)\) and associated labels \(y \in \{0, 1\}\) indicating whether the sequences are paraphrases or not. Because the Opusparcus data contains ranked paraphrase candidates and not labeled pairs, we take the following approach to sampling the data: The desired number of paraphrase pairs (positive examples) are taken from the beginning of the data sets. That is, the highest ranking pairs, which are the most likely to be proper paraphrases according to \newcite{creutz2018lrec}, are labeled as paraphrases, although not all of them are true paraphrases. The non-paraphrase pairs (negative examples) are created by randomly pairing sentences from the training data. It is possible that a positive example is created this way by accident, but we assume the likelihood of this to be low enough for it not to have noticeable effect on performance. We sample an equal number of positive and negative pairs in all experiments. In the rest of this paper, when mentioning training set sizes, we indicate the number of (assumed) positive pairs sampled from the data. There is always an equal amount of (assumed) negative pairs.

During training we optimize the following margin-based loss function:
\begin{equation*}\begin{split}
L(\theta) =~&y(max(0, m - d(g(s_1), g(s_2)))^2 \\ &+ (1 - y)d(g(s_1), g(s_2))
\end{split}\end{equation*}

\noindent Here \(m\) is the margin parameter, \(d(g(s_1), g(s_2))\) is the cosine distance between the embedded sequences, and \(g\) is the embedding function. The loss function penalizes negative pairs with a cosine distance smaller than the margin (first term) and encourages positive pairs to be close to each other (second term).

We use the Adam optimizer \cite{kinga2015iclr} with a learning rate of 0.001 and a batch size of 128 samples in all experiments. Variational dropout \cite{gal2016nips} is used for regularization in the GRAN model. The hyperparameters were tuned in preliminary experiments for development set accuracy and, with the exception of keep probability in dropout, kept constant in all experiments.

The embedding matrix \(W\) is initialized to a uniform distribution over \([-0.01, 0.01]\). In our experiments we found that initializing with pre-trained embeddings did not improve the paraphrase detection results. The layer weights in the GRU network are initialized using Xavier initialization \cite{glorot2010aistats}, and we use the leaky ReLU activation function.

\section{Experiments}

Our initial experiment addresses the effects of unsupervised morphological segmentation on the results of the paraphrase detection task.

Next, we tackle our main question on the trade-off between the amount of noise in the training data and the data size. In particular, we try to see if an optimal amount of noise can be found, and whether the different models have different demands in this respect.

 Finally, we evaluate the English-language models on out-of-domain semantic similarity and paraphrase detection tasks.

 All evaluations on the Opusparcus are conducted in the following manner: Each sentence in the sentence pair is embedded using the sentence encoding model. The resulting vectors are concatenated and passed on to a multi-layer perceptron classifier with a single hidden layer of 200 units. The classifier is trained on the development set, and the final results are reported on the unseen test set in terms of classification accuracy.

\subsection{Segmentation}
\label{sec:segmentation}

We work on six different European languages, some of which are morphologically rich (that is, the number of possible word forms in the language is high). In the case of languages like Finnish and Russian, the vocabularies without any kind of morphological preprocessing can grow very large even with small amounts of data.

In our approach we train Morfessor Baseline \cite{creutz2002acl,virpioja2013tr}, an unsupervised morphological segmentation algorithm, on the whole Opusparcus training data available. Segmentation approaches that result in fixed-size vocabularies, such as byte-pair encoding (BPE) \cite{sennrich2015acl}, have been gaining popularity in some natural language processing tasks. We decided to use Morfessor instead, which also appeared to outperform BPE in preliminary experiments. However, we will not focus on segmentation quality, but use segmentation simply as a preprocessing step to improve downstream performance.

The results are shown in the WA-M and WA columns of Table \ref{tab:2mresults}. The differences in performance between the WA models with segmentation (called just WA) and without segmentation (called WA-M) clearly indicate that this is a necessary preprocessing step when working on languages with complex morphology. The effect of segmentation for the GRAN model (not shown) is similar, with the exception of English also improving by a few points instead of worsening. Based on these results we will use Morfessor as a preprocessing step in all of the remaining experiments.

\begin{table}
  \begin{center}
    \begin{tabular}{| c | c c c c |}
        \hline
         & AP   & WA-M   & WA & GRAN \\ \hline
      de & 74.3 & 77.0 & 82.3 & 83.2 \\
      en & 72.8 & 87.4 & 86.4 & 89.2 \\
      fi & 61.0 & 74.7 & 80.3 & 80.1 \\
      fr & 68.6 & 74.0 & 76.7 & 76.8 \\
      ru & 65.4 & 61.4 & 70.9 & 69.7 \\
      sv & 54.8 & 78.1 & 84.1 & 83.2 \\ \hline
    \end{tabular}
  \end{center}
  \caption{\label{tab:2mresults}Classification accuracies on the Opusparcus test sets for models trained on 1 million positive sentence pairs. AP (all paraphrases) is the majority baseline, which is the accuracy obtained if all sentence pairs in the test data are labeled as paraphrases. Consistent improvement is obtained by the WA model without segmentation (WA-M: ``WA without Morfessor'') and further by the WA model with segmentation. Whether the GRAN model outperforms WA is hard to tell from these figures, but this is further analyzed in Section~\ref{sec:selection_experiments}.}
\end{table}

\subsection{Data selection}
\label{sec:selection_experiments}

We next investigate the effects of data set size and the amount of noise in the data on model performance. We are interested in finding an appropriate amount of training data to be used in training the paraphrase detection models, as well as evaluating the robustness of different models against noise in the data.

For each language, data sets containing approximately 80\%, 70\%, or 60\% clean paraphrase pairs are created. These percentages are the proportions of assumed positive training examples; the negative examples are created using the approach outlined in Section~\ref{sec:training}.

Estimates of the quality of the training sets exist for all languages in Opusparcus.\footnote{The figures used to approximate the data set sizes can be found in the presentation slides (slides 12-13) at \url{https://helda.helsinki.fi//bitstream/handle/10138/237338/creutz2018lrec_slides.pdf}} The quality estimates were used to approximate the numbers of phrase pairs corresponding to the noise levels. Because the data sets for different languages are not equal in size, the number of phrase pairs at a certain noise level differs from language to language. The different data set sizes for all noise levels and languages are shown in Table~\ref{tab:noiseresults}.



\begin{table*}
  \begin{center}
    \begin{tabular}{| c | l || l l l |}
        \hline
         &   1M        & 80\%               & 70\%                & 60\%               \\ \hline
      de & 83.2 (90\%) & \textbf{86.7 (4)}  & 85.3 (6)            & 85.6 (12)          \\
      en & 89.2 (97\%) & 90.2 (5)           & \textbf{92.1 (20)}  & 90.9 (34)          \\
      fi & 80.1 (83\%) & 81.4 (2.5)         & \textbf{82.5 (3.5)} & 81.5 (9)           \\
      fr & 76.8 (95\%) & 76.2 (5)           & 77.1 (13)           & \textbf{77.9 (22)} \\
      ru & 69.7 (85\%) & 60.3 (2)           & \textbf{70.3 (5)}   & 66.8 (15)          \\
      sv & 83.2 (85\%) & 71.7 (1.2)         & 73.0 (1.8)          & \textbf{82.1 (5)}  \\ \hline
      en (WA) & \textbf{86.4 (97\%)} & 79.5 (5)      & 77.9 (20)           & 77.2 (34)          \\ \hline
    \end{tabular}
  \end{center}
  \caption{\label{tab:noiseresults}Results on Opusparcus for GRAN (all languages) and WA (English only). The first six rows show the accuracies of the GRAN model at different estimated levels of correctly labeled positive training pairs: 80\%, 70\%, and 60\%. In each entry in the table, the first number is the classification accuracy and the number in brackets is the number of assumed positive training pairs in millions. For comparison, the 1M column to the left repeats the values from Table~\ref{tab:2mresults}, in which the size of the training set was the same for each language, regardless of noise levels; the estimated proportion of truly positive pairs in these setups are shown within brackets. The last row of the Table shows the performance of the WA model for English.}
\end{table*}

Table~\ref{tab:noiseresults} shows the results for the GRAN model. The results indicate that the GRAN model is rather robust to noise in the data. For five out of six languages, the best results are achieved using either the 70\% or 60\% data sets. That is, even when up to 40\% of the positive examples in the training data are incorrectly labeled, the model is able to maintain or improve its performance.

The results for the WA model are very different. The last row of Table \ref{tab:noiseresults} shows the accuracies of the WA model at different levels of noise for English. The model's performance decreases significantly as the number of noisy pairs increases, and the results are similar for the other languages as well. We hypothesize these differences to be due to differences in model complexity. The GRAN model incorporates a sequence model and contains more parameters than the simpler WA model.

\subsubsection{Further analysis of differences between models}
\label{sec:further_analysis}

\begin{figure*}[!t]
  \begin{center}
    \includegraphics[width=.5\textwidth]{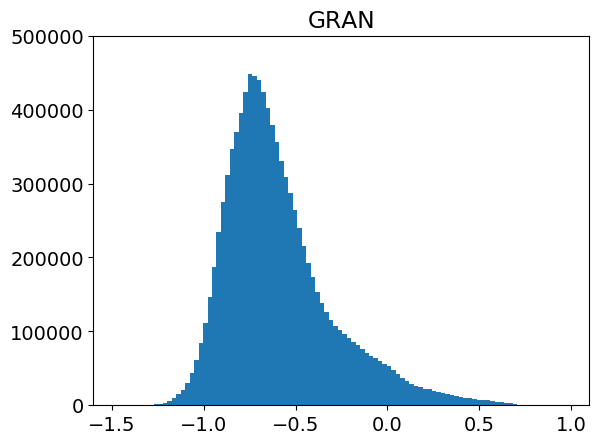}\includegraphics[width=.5\textwidth]{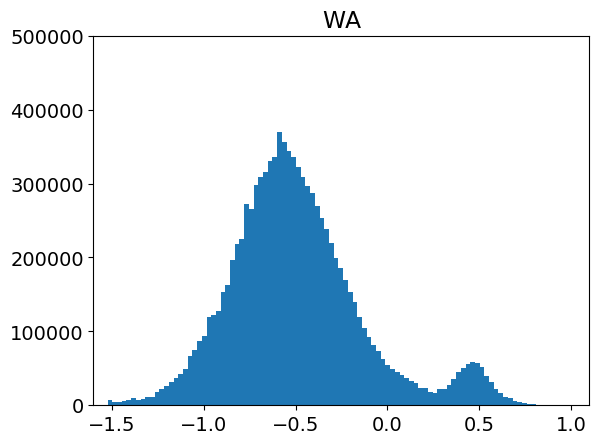}
    \caption{Distributions of similarity scores between the target sentence ``\textit{okay, you don't get it, man.}'' and 10~million English sentences from OpenSubtitles. Cosine similarity between sentence embedding vectors are used. A sentence that is very close to the target sentence has a cosine similarity close to 1, whereas a very dissimilar sentence has a value close to -1. (Some of the similarity values are below -1 because of rounding errors in Faiss: {\footnotesize \url{https://github.com/facebookresearch/faiss/issues/297}}.) Section~\ref{sec:further_analysis} discusses differences in the distributions between the GRAN and WA models.
}
    \label{fig:similaritydistributions}
  \end{center}
\end{figure*}

Some qualitative differences between the WA and GRAN models are illustrated in Tables~\ref{tab:you_dont_get_it} and \ref{tab:kann_gut_sein} as well as Figure~\ref{fig:similaritydistributions}. Table~\ref{tab:you_dont_get_it} shows which ten sentences in the English development set are closest to one target sentence ``\textit{okay, you don't get it, man.}'' according to the two models. The comparison is performed by computing the  cosine similarity between the sentence embedding vectors. A similar example is shown for German in Table~\ref{tab:kann_gut_sein}: ``\textit{Kann gut sein.}'' (in English: ``\textit{That may be.}'')\footnote{Further examples of similar sentences can be found in the supplemental material.}

The result suggests that the WA (word averaging) models produce ``bag of synonyms''. Sentences are considered similar if they contain the same words or similar words. This, however, makes the WA model perform weakly when a sentence should not be interpreted literally word by word. German ``\textit{Kann gut sein.}'' is unlikely to literally mean ``\textit{Can be good.}''; yet sentences with that meaning are at the top of the WA ranking. By contrast, the GRAN model comes up with very different top candidates, sentences expressing modality, such as: ``\textit{Possibly}'', ``\textit{Yes, he might}'', ``\textit{You're probably right}'', ``\textit{As naturally as possible}'', and ``\textit{I think so}''.

Figure~\ref{fig:similaritydistributions} provides some additional information on the English sentence ``\textit{okay, you don't get it, man.}''. Distributions of the cosine similarities of a much larger number of sentences have been plotted (10 million sentences from English OpenSubtitles). In the plots, similar sentences are on the right and dissimilar sentences on the left. In the case of the GRAN model we see a huge mass of dissimilar sentences smoothing out in a tail of similar sentences. In the case of the WA model, there is clearly a second, smaller bump to the right. It turns out that the ``bump'' mainly contains negated sentences, that is, sentences that contain synonyms of ``\textit{don't}''. A second look at Table~\ref{tab:you_dont_get_it} validates this observation: the common trait of the sentences ranked at the top by WA is that they contain ``\textit{don't}'' or ``\textit{not}''. Thus, according to WA, the main criterion for a sentence to be similar to ``\textit{okay, you don't get it, man.}'' is that the sentence needs to contain negation. Again, the GRAN model stresses other, more relevant aspects, in this case, whether the sentence refers to \textit{not knowing} or \textit{not understanding}.

\begin{table}[!t]
\begin{center}
\begin{tabular}{|c|lr|}
\hline
& okay , you don 't get it , man . & \\
\hline
& \textbf{you don 't understand .} & {\small 0.98 } \\
& no , you don 't understand . & {\small 0.98 } \\
& you can 't know that . & {\small 0.92 } \\
G & you do not really know . & {\small 0.90 } \\
R & no , i don 't think you understand & {\small 0.88 } \\
A & you know , nobody has to know . & {\small 0.86 } \\
N & you don 't got it . & {\small 0.82 } \\
& no one will ever know . & {\small 0.82 } \\
& and no one will know . & {\small 0.81 } \\
& we don 't know yet . & {\small 0.81 } \\
\hline
& you don 't got it . & {\small 0.91 } \\
& don 't go over . & {\small 0.91 } \\
& do not beat yourself up about that . & {\small 0.90 } \\
& please don 't . & {\small 0.89 } \\
W & well ... not everything . & {\small 0.89 } \\
A & not all of it . & {\small 0.88 } \\
& you don 't have to . & {\small 0.87 } \\
& no , you don 't understand . & {\small 0.87 } \\
& one it 's not up to you . & {\small 0.86 } \\
& okay , that 's not necessary . & {\small 0.84 } \\
\hline
\end{tabular}
\end{center}
\caption{\label{tab:you_dont_get_it} The ten most similar sentences to ``\textit{okay, you don't get it, man.}'' in the Opusparcus English development set, based on sentence embeddings produced by the GRAN and WA models, respectively. Cosine similarities are shown along with the sentences. (The annotated ``correct'' paraphrase is ``\textit{you don't understand.}'')  }
\end{table}

\begin{table}[!t]
\begin{center}
\begin{tabular}{|c|lr|}
\hline
& Kann gut sein . & \\
\hline
& M\"oglicherweise . & {\small 0.93 } \\
& Ja , k\"onnte er . & {\small 0.92 } \\
& Hast wohl Recht . & {\small 0.92 } \\
G & So nat\"urlich wie m\"oglich . & {\small 0.91 } \\
R & Ihr habt nat\"urlich recht . & {\small 0.91 } \\
A & Sie haben recht , nat\"urlich . & {\small 0.88 } \\
N & Ich denke , doch . & {\small 0.88 } \\
& Ja , ich denke schon . & {\small 0.87 } \\
& \textbf{Wahrscheinlich schon .} & {\small 0.87 } \\
& Ich bin mir sicher . & {\small 0.87 } \\
\hline
& Das ist doch gut . & {\small 0.83 } \\
& Na , das ist gut . & {\small 0.81 } \\
& Ist in Ordnung . & {\small 0.81 } \\
& Dir geht es gut . & {\small 0.81 } \\
W & Ihnen geht es gut . & {\small 0.81 } \\
A & Sie ist in Ordnung . & {\small 0.81 } \\
& Ich kann es f\"uhlen . & {\small 0.80 } \\
& Es ist alles gut . & {\small 0.79 } \\
& Mir geht 's gut . & {\small 0.79 } \\
& Sie is okay . & {\small 0.79 } \\
\hline
\end{tabular}
\end{center}
\caption{\label{tab:kann_gut_sein} The ten most similar sentences to ``\textit{Kann gut sein.}'' in the Opusparcus German development set, based on sentence embeddings produced by the GRAN and WA models, respectively. The annotated ``correct'' paraphrase is here ``\textit{Wahrscheinlich schon.}'' (``\textit{Probably yes}'').}
\end{table}

%
%
%
%

\subsection{PPDB as training data}
\label{sec:ppdb}

We also train the GRAN model on PPDB data. \citet{wieting2017acl} found that models trained on PPDB achieve good results on a wide range of semantic textual similarity tasks, thus, good performance could be expected on the Opusparcus test sets.

For English we use the PPDB 2.0 release, for languages other than English we use the 1.0 release, as the 2.0 is not available for those languages. The phrasal paraphrase packs are used for all languages. We pick the number of paraphrase pairs in such a way that the training data contains as close to an equal number of tokens as the Opusparcus training data with 1 million positive examples. This ensures that the amount of training data is as similar as possible in both settings. The training setup is otherwise identical to that outlined above.

The results are shown in Table \ref{tab:ppdbresults}. There is a significant drop in performance when moving from in-domain training data (Opusparcus) to out-of-domain training data (PPDB). One possible explanation for this is that the majority of the phrase pairs in the PPDB dataset contain sentence fragments rather than full sentences.


\begin{table}
  \begin{center}
    \begin{tabular}{| c | c |}
        \hline
         & GRAN \\ \hline
      de & 78.1 \\
      en & 83.4 \\
      fi & 70.4 \\
      fr & 74.8 \\
      ru & 67.7 \\
      sv & 76.4 \\ \hline
    \end{tabular}
  \end{center}
  \caption{\label{tab:ppdbresults}Results on Opusparcus test sets for models trained on PPDB.}
\end{table}

\subsection{Transfer learning}
\label{sec:transfer}

We also evaluate our English models on other data sets. Because we are primarily interested in paraphrastic sentence embeddings, we choose to evaluate our models on the MSRPC paraphrase corpus, as well as two semantic textual similarity tasks, SICK-R and STS14. The data represent a range of genres, and hence offer a view of the potential of Opusparcus for out-of-domain use and transfer learning. Because of the similarities between paraphrase detection and the semantic textual similarity tasks, we believe the two tasks to be mutually supportive.

We present results for the WA model as well as the best GRAN model from Section \ref{sec:selection_experiments}. The evaluations are conducted using the SentEval toolkit \cite{conneau2018senteval}. To obtain comparable results, we use the recommended default configuration for the SentEval parameters. The results are shown in Table \ref{tab:transferresults}.

We first note that our models fall short of the state-of-the-art results by a rather large margin. We hypothesize the discrepancy between the performance on MSRPC of our models and the BiLSTM-Max model of \citet{conneau2017acl} to be due to differences in the genre of training data. The conversational language of subtitles is vastly different from the news domain of MSRPC. Although the NLI data used by \citet{conneau2017acl} is derived from an image-captioning task and thus does not represent the news domain, it is at least closer to MSRPC in terms of the vocabulary and sentence structure. Most interesting is the difference between our WA model and the Paragram-phrase model of \citet{wieting2016iclr}. These are essentially the same model, but trained on two different data sets. While the performance on SICK-R is comparable, our model significantly underperforms on STS14. Overall the results indicate that our models tend to overfit the domain of the Opusparcus data and consequently do not perform as well on out-of-domain data.

\begin{table}
  \begin{center}
    \begin{tabular}{| l | c c c |}
        \hline
                           & MSRPC      & SICK-R &    STS14  \\ \hline
           GRAN            & 69.5/80.6 &   .717 &   .40/.44 \\
           WA              & 67.1/79.1 &   .710 &   .54/.53 \\ \hline \hline
           BiLSTM-Max      & 76.2/83.1 &   .884 &   .70/.67 \\
           Paragram-phrase & -         &   .716 &   .71/-   \\
           FastSent        & 72.2/80.3 &      - &   .63/.64 \\
      \hline
    \end{tabular}
  \end{center}
  \caption{\label{tab:transferresults}Transfer learning results on MSRPC, SICK-R and STS14. GRAN and WA denote our models. We also show results for a selection of models from the transfer learning literature. We use the evaluation measures that are customarily used in connection with these data sets. For MSRPC, the accuracy (left) and F1-score (right) are reported. For SICK-R we report Pearson's $r$, and for STS14 Pearson's $r$ (left) and Spearman's rho (right). For all these measures a higher value indicates a better result.}
\end{table}

\section{Discussion and Conclusion}

Our results show that even a large amount of noise in training data is not always detrimental to model performance. This is a promising result, as automatically collected, large but noisy data sets are often easier to come by than clean, manually collected or annotated data sets. Our results can also guide model selection when noise in training data is a consideration.

In future work we would like to explore how to most effectively leverage possibly noisy paraphrase data in learning general-purpose sentence embeddings for a wide range of transfer tasks. Investigating training procedures and encoding architectures that allow for robust models with the capability for generalization is a topic for future research.

%

\bibliography{wnut2018}
\bibliographystyle{acl_natbib_nourl}

\clearpage

\appendix

\section{Supplemental Material}
\label{sec:supplemental}

The following sections present some additional and clarifying results that validate choices that were made in our experiments.

\subsection{Unsupervised morphological segmentation}

\begin{table}
  \begin{center}
     \begin{tabular}{| c | c c |}
         \hline
          & No segmentation & Morfessor \\ \hline
       de &           65437 &     12329 \\
       en &           56571 &     15295 \\
       fi &          130879 &     33513 \\
       fr &           69920 &     18143 \\
       ru &          137942 &     55480 \\
       sv &           81407 &     16397 \\ \hline
     \end{tabular}
   \end{center}
   \caption{\label{tab:vocabs}Vocabulary sizes for 2 million phrase pairs from Opusparcus. The values contain both positive and negative examples used in training.}
 \end{table}

Table \ref{tab:vocabs} shows the vocabulary sizes for the training data for the six languages. The training data sets each consist of two million paraphrase pairs. The Morfessor Baseline algorithm is utilized to split words into smaller subword units, partly resembling morphemes. The segmentation of words into such subword units is clearly effective in reducing the vocabulary size for all languages. As expected, particularly drastic reductions can be seen for Finnish and Russian as they are the most morphologically complex languages of the six.

\subsection{Effect of Morfessor on GRAN}

\begin{table}
  \begin{center}
    \begin{tabular}{| c | c c |}
        \hline
         & GRAN-M & GRAN \\ \hline
      de & 78.2   & 83.2 \\
      en & 87.6   & 89.2 \\
      fi & 72.8   & 80.1 \\
      fr & 74.9   & 76.8 \\
      ru & 67.1   & 69.7 \\
      sv & 79.4   & 83.2 \\ \hline
    \end{tabular}
  \end{center}
  \caption{\label{tab:morf_gran}Results on Opusparcus test sets for GRAN models trained on 1 million positive sentence pairs. Shown is classification accuracy. -M indicates a model trained without Morfessor.}
\end{table}

Table~\ref{tab:morf_gran} shows the effect of unsupervised morphological segmentation on the GRAN model. The differences show the usefulness of the Morfessor segmentation as a preprocessing step as discussed in Section~\ref{sec:segmentation} of the main paper. The main difference between the GRAN and WA models is the direction of change for English. The English WA model works better without morphological segmentation, whereas the performance of the GRAN model is clearly improved.

\subsection{Further examples of similar sentences}

Tables~\ref{tab:were_on_a_clock} and \ref{tab:je_creve} show additional examples of sentences that are most similar to a target sentence according to the GRAN and WA models. These examples are along the lines of the discussion in Section~\ref{sec:further_analysis} and Tables~\ref{tab:you_dont_get_it} and \ref{tab:kann_gut_sein} of the main paper. The GRAN model favors sentences that actually relate to the topic of the target sentence (\textit{running out of time} in the English examples of Table~\ref{tab:were_on_a_clock}, and \textit{dying} or \textit{surviving} in the French examples of Table~\ref{tab:je_creve}). The WA model promotes synonyms of individual words, and when these words are common function words, such as pronouns, prepositions, and  common verbs, the meaning of the suggested sentence can be completely off topic.

\begin{table}[!t]
\begin{center}
\begin{tabular}{|c|lr|}
\hline
& we 're on a clock . & \\
\hline
& \textbf{the clock is ticking .} & {\small 0.94 } \\
& it 's time now . & {\small 0.91 } \\
& in the mean time . & {\small 0.90 } \\
G & time is running out , doctor . & {\small 0.89 } \\
R & they took their time . & {\small 0.89 } \\
A & we 're out of time . & {\small 0.89 } \\
N & there will not be another time . & {\small 0.88 } \\
& ain 't no next time . & {\small 0.86 } \\
& playing for time . & {\small 0.86 } \\
& this is not a good time , okay ? & {\small 0.84 } \\
\hline
& we 're out of time . & {\small 0.87 } \\
& we 're together . & {\small 0.83 } \\
& we 've been through this . & {\small 0.80 } \\
& \textbf{the clock is ticking .} & {\small 0.80 } \\
W & we 've done this before . & {\small 0.79 } \\
A & catch you next time . & {\small 0.77 } \\
& sometimes that 's all you need . & {\small 0.77 } \\
& we 've been over this , okay . & {\small 0.77 } \\
& they took their time . & {\small 0.77 } \\
& we have an assignment . & {\small 0.77 } \\
\hline
\end{tabular}
\end{center}
\caption{\label{tab:were_on_a_clock} The ten most similar sentences to ``\textit{we're on a clock.}'' in the Opusparcus English development set, based on sentence embeddings produced by the GRAN and WA models, respectively. Cosine similarities are shown along with the sentences. (The annotated ``correct'' paraphrase is ``\textit{the clock is ticking.}'')}
\end{table}

\begin{table}[!t]
\begin{center}
\begin{tabular}{|c|lr|}
\hline
& tu veux que je cr\`eve ? & \\
\hline
& est - elle en train de mourir ? & {\small 0.99} \\
& tu a s d\'ej\`a vu un homme mourir ? & {\small 0.98} \\
& elle est mourante ? & {\small 0.98} \\
G & {\fontsize{10}{12}\selectfont que peut vouloir de plus une mourante ?}  & {\small 0.98} \\
R & \c{c}a fait dispara\^itre mes pouvoirs ? & {\small 0.98} \\
A & suis - je le seul survivant ? & {\small 0.98} \\
N & vous pensez qu' on vous a pi\'eg\'e ? & {\small 0.97} \\
& c' est \c{c}a qui a provoqu\'e sa mort ? & {\small 0.97} \\
& on l' a sem\'e ? & {\small 0.97} \\
& c' est \`a cause de \c{c}a qu' il est mort ? & {\small 0.97} \\
\hline
& que puis - je faire pour toi ? & {\small 0.94} \\
& je peux venir avec vous ?  & {\small 0.92} \\
& puis - je t' accompagner ?  & {\small 0.92} \\
& {\fontsize{10}{12}\selectfont est - ce que tu me fais une proposition ?} & {\small 0.91} \\
W &  je vous d\'epose ? & {\small 0.91} \\
A & qu' est - ce que tu me veux ? & {\small 0.91} \\
& est ce que je t' ai d\'ej\`a d\'e\c{c}ue ? & {\small 0.90} \\
& \c{c}a fait dispara\^itre mes pouvoirs ? & {\small 0.90} \\
& c' est ce que tu d\'esires ? & {\small 0.90} \\
& mais qu' est - ce que tu me veux ? & {\small 0.90} \\
\hline
\end{tabular}
\end{center}
\caption{\label{tab:je_creve} The ten most similar sentences to ``\textit{tu veux que je cr\`eve~?}'' (``\textit{do you want me to die?}'') in the Opusparcus French  development set, based on sentence embeddings produced by the GRAN and WA models, respectively. Cosine similarities are shown along with the sentences. (There is no true paraphrase for this target sentence in the data, as the sentence proposed tentatively as a paraphrase is ``\textit{t'aurais ma mort sur la conscience}'', which means``\textit{you'd have my life on your conscience}'', which the annotators disqualified as a proper paraphrase.)}
\end{table}

\subsection{Validation of automatically assigned similarity scores}

The Opusparcus development and test sets contain sentence pairs accompanied by categories assigned by human annotators. Each annotator used a four-grade scale: \textit{dark green} (good), \textit{light green} (mostly good), \textit{yellow} (mostly bad), or \textit{red} (bad). This four-grade scale can be extended to a seven-grade scale, if we add extra categories in between the given ones: if both annotators that saw a particular sentence pair agreed on a category, such as \textit{yellow} or \textit{red}, the final appropriate category is clear, but if the annotators chose adjacent categories, such as \textit{yellow} and \textit{red}, we can insert an additional category, in this case \textit{orange}, between \textit{yellow} and \textit{red}.

\begin{figure*}[h]
\begin{center}
\includegraphics[width=1.04\textwidth]{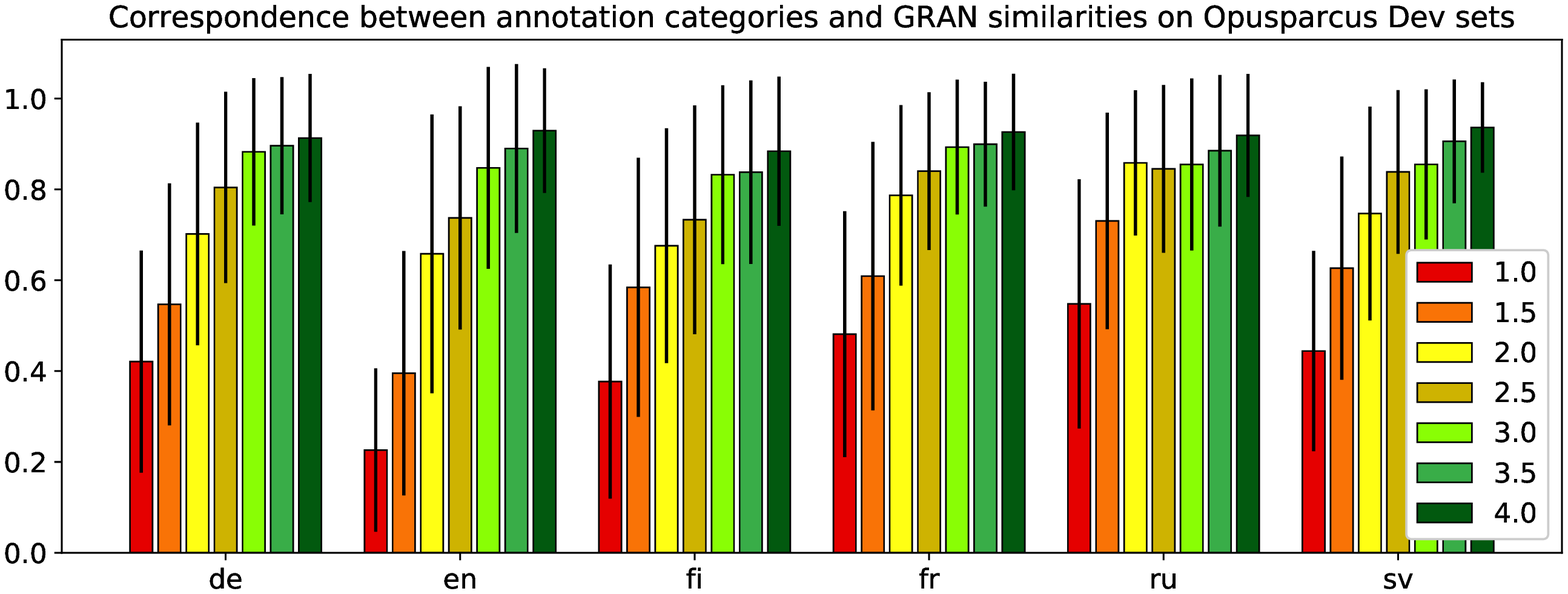}
\includegraphics[width=1.04\textwidth]{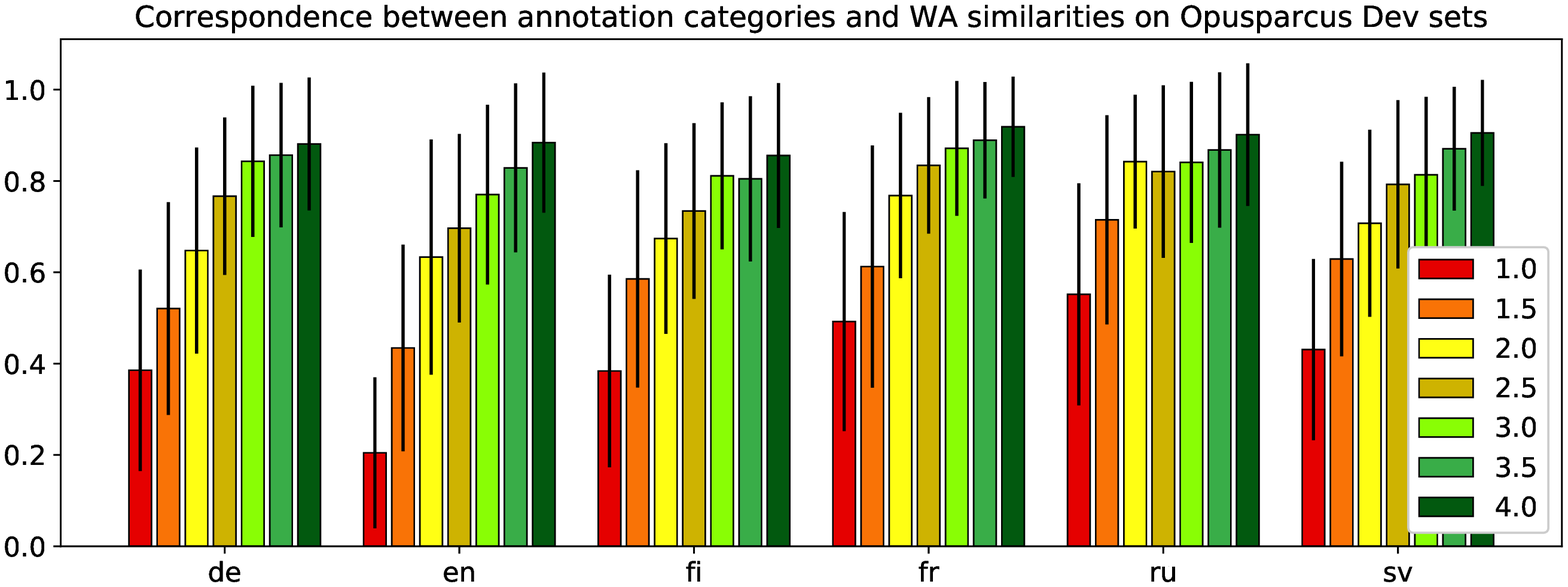}
\caption{Comparison of human-assigned and automatically determined similarity scores. The upper plot refers to the GRAN model and the lower plot to the WA model. For each model and language, there are seven vertical bars. The height of a bar represents the mean cosine similarity of all sentence pairs that have the same human-assigned score in the range $[1.0, 4.0]$. The black vertical lines show the standard deviation. The plots illustrate that the higher the human-assigned score is, the higher cosine similarity, in general. Additionally, the differences in cosine similarity are mostly statistically significant between the steps, except the following for GRAN: 4.0 vs. 3.5, 3.5 vs. 3.0 (de), 2.5 vs 2.0 (en), 3.5 vs 3.0, 2.5 vs. 2.0 (fi), 3.5 vs. 3.0 (fr), 3.5 vs. 3.0, 3.0 vs. 2.5, 2.5 vs 2.0 (ru), 3.0 vs. 2.5 (sv). Exactly the same comparisons are statistically significant for WA, but unlike GRAN, the difference between 2.5 vs. 2.0 (fi) is here statistically significant. T-tests were performed using the 0.01 significance level.}
\label{fig:similaritybars}
\end{center}
\end{figure*}

The automatic paraphrase detection models that we train (GRAN and WA) produce sentence embedding vectors. These vectors can be compared using, for instance, cosine similarity. For each of the six languages in the Opusparcus data, we decided to compare the seven-grade scores assigned manually by humans to cosine similarity scores obtained automatically from the GRAN and WA models. We used the development sets in our comparison and the results are shown in Figure~\ref{fig:similaritybars}.

Ideally, we would like to see that low scores assigned by humans, such as 1.0 (\textit{red}), 1.5 (\textit{orange}), and 2.0 (\textit{yellow}), correspond to low cosine similarities, and that high scores, such as 4.0 (\textit{dark green}), 3.5 (\textit{medium green}), and 3.0 (\textit{light green}), correspond yo high cosine similarities. In general, this seems to be the case. For both models (GRAN and WA) and for every language, except Russian between 2.0 (\textit{yellow}) and 3.0 (\textit{light green}), the higher the human-assigned score, the higher the automatically determined cosine similarity, on average. Most of the differences are also statistically significant according to T-tests at the 0.01 significance level. These observations suggest that human judgment and the automatic scores produced by the WA and GRAN models are generally in agreement, although not always for each and every sentence pair.

\end{document}